\documentclass[10pt,twocolumn,a4paper]{article}

\usepackage[T1]{fontenc}
\usepackage{lmodern}
\usepackage{microtype}
\usepackage[a4paper,top=18mm,bottom=20mm,left=17mm,right=17mm,columnsep=6mm]{geometry}
\usepackage{graphicx}
\usepackage{booktabs}
\usepackage{array}
\usepackage{amsmath}
\usepackage{siunitx}
\usepackage{xcolor}
\usepackage{xurl}
\usepackage{hyperref}
\usepackage{cleveref}
\usepackage{caption}
\usepackage{subcaption}
\usepackage{enumitem}
\usepackage{balance}
\usepackage{tikz}
\usetikzlibrary{arrows.meta,positioning,fit,calc}

\definecolor{gybeblue}{HTML}{126E82}
\definecolor{gybeorange}{HTML}{E67E22}
\definecolor{gybered}{HTML}{C0392B}
\newcolumntype{L}[1]{>{\raggedright\arraybackslash}p{#1}}
\setlist{nosep,leftmargin=*}
\hypersetup{
  colorlinks=true,
  linkcolor=gybeblue,
  urlcolor=gybeblue,
  citecolor=gybeblue,
  pdftitle={Identity-Consistent Analysis of Long-Shot Windsurfing Video},
  pdfauthor={Bertil Braun; Felix M\"uhlenberend}
}

\graphicspath{{figures/}{../}{../../frontend/public/}}

\title{\vspace{-8mm}\textbf{Identity-Consistent Analysis of Long-Shot Windsurfing Video}\\
\large A Domain-Specific Offline Tracking System}
\author{%
\begin{tabular}{c@{\hspace{3em}}c}
Bertil Braun & Felix M\"uhlenberend \\
\href{mailto:contact@bertil-braun.de}{contact@bertil-braun.de} &
\href{mailto:felix.muehlenberend@mailbox.org}{felix.muehlenberend@mailbox.org}
\end{tabular}}
\date{}

\begin{document}
\twocolumn[
\begin{@twocolumnfalse}
\maketitle
\vspace{-5mm}

\begin{abstract}
Long-shot windsurfing video combines small targets, large camera pans, prolonged
overlaps, and rapidly changing backgrounds.  The desired output is not a generic MOT
trace but a separate, stable rider-relative video for each surfer; one false identity
merge can invalidate an otherwise useful result.  We present an offline analysis
system that detects surfers, forms conservative local tracklets, links them globally
with camera-compensated motion and a foreground-masked sail-color descriptor, and
uses two pose keypoints on the rig to drive a rider-relative virtual camera.  The tracking stage is
evaluated on 21 manually reconstructed development videos containing 41,004 retained
observations.  On this fixed-observation protocol, the production system achieves
0.957 pairwise precision, 0.918 recall, and 0.937 F1, compared with 0.792 F1 for
OC-SORT and 0.828 for BoT-SORT.  Compared with OC-SORT, it reduces fragmentation
excess from 845 to 42, but nine of its 95 output tracks mix rider identities and these
errors affect seven of the 21 videos.
\end{abstract}
\vspace{0.5em}
\vspace{1em}
\end{@twocolumnfalse}
]

\section{Introduction}

Shore-recorded windsurfing video is visually rich but difficult to repurpose into a
focused view of one athlete.  The camera follows the action with large pans, riders
occupy few pixels, rigs overlap for extended periods, and an athlete may disappear
behind another sail before returning against a different background.  A useful system
must do more than draw plausible boxes: it must preserve one physical identity long
enough to produce a stable rider-relative video.

This requirement changes the engineering objective.  A missed association leaves two
fragments that can be reviewed or joined later.  A false association silently inserts
another surfer into the output and can invalidate the complete sequence.  Standard
online trackers are designed for causal decisions and are typically evaluated with
metrics that balance detection and association errors.  Here both failure modes
matter but not equally: a false merge can corrupt a complete rider video, whereas
excessive fragmentation can leave even pure tracks unusable.  The full recording is
available, so the system favors conservative local decisions and recovers continuity
later through global association.

We describe the resulting end-to-end computer-vision system.  A custom-trained
pose detector supplies surfer boxes and two rig keypoints.  Masked global-motion estimation
separates camera pan from target motion.  Conservative online association creates
high-purity tracklets, after which an offline integer program links compatible
tracklets using motion, temporal gap, and a sail-specific color representation.
Finally, the pose keypoints define a temporally smoothed anchor and scale for a stable
virtual camera.  Source code, technical documentation, and an output demonstration
are publicly available for inspection~\cite{projectresources}.

The contributions are:
\begin{itemize}
  \item an explicit formulation of long-shot windsurfing analysis around asymmetric
  costs for identity merges and fragmentation, with complete-video processing;
  \item a domain-specific association pipeline combining foreground-masked sail
  appearance, camera-compensated four-dimensional motion, conservative tracklets,
  geometric vetoes, and global path optimization;
  \item a two-keypoint rider-relative stabilization signal designed around windsurf
  rig geometry rather than generic human pose; and
  \item a fixed-observation evaluation on 21 manually reconstructed videos, including
  OC-SORT and BoT-SORT comparisons, staged ablations, and a link-level failure audit.
\end{itemize}

\section{Problem Regime and Requirements}
\label{sec:problem}

The input is a complete windsurfing recording captured from shore.  A scene usually
contains only a handful of surfers, but each target can be small and several sails may
remain close or overlap for seconds.  The camera follows the action with pans of about
\SI{90}{\degree} in demanding examples, sometimes more.  Consequently almost every
background feature moves in image coordinates, and newly revealed portions of the
scene may share no features with earlier frames.

\begin{figure*}[t]
  \centering
  \includegraphics[width=0.96\textwidth]{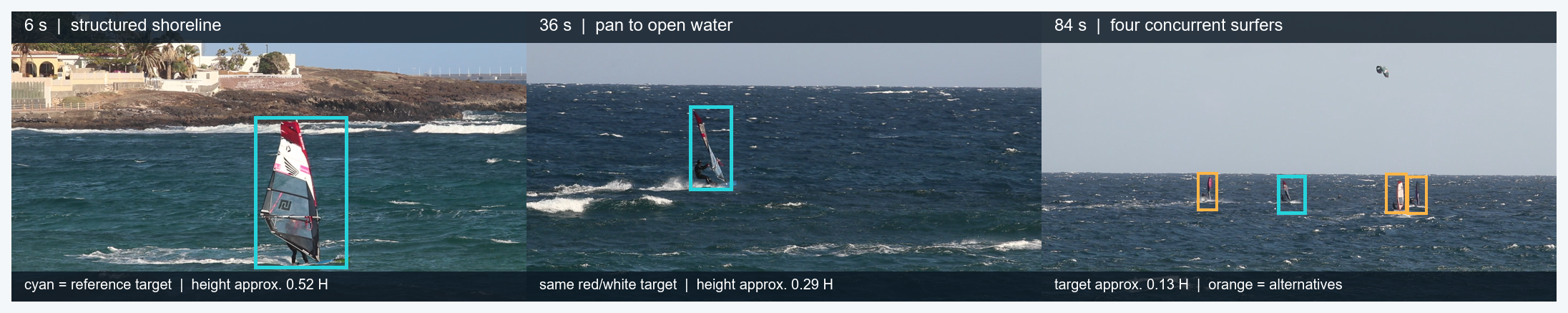}
  \caption{Operating regime in one recording at 6, 36, and \SI{84}{\second}.  Cyan
  follows the same red-and-white sail while the background and camera direction
  change; the target height drops from approximately $0.52H$ to $0.13H$.  Orange
  marks plausible competing surfers.  The example combines scale change, camera
  motion, and identity ambiguity within one continuous shot.}
  \label{fig:problem-operating-regime}
\end{figure*}

The desired result is one continuous, dense trajectory per physical surfer from which
a focused video can be rendered.  This gives three requirements.

\paragraph{Identity is the primary invariant.}
A track that switches rider once is not partly correct from the user's perspective:
the extracted sequence mixes identities.  Localization smoothness and recall matter
only after identity is preserved.

\paragraph{Association errors are asymmetric.}
A missed link creates fragments that remain inspectable.  A false link combines two
identities and is difficult to detect downstream.  The system should therefore
abstain when evidence is insufficient.  ``Near-zero false merges'' is the design
target, not a claim that the implementation achieves perfect tracking.

\paragraph{The complete recording is available.}
There is no real-time constraint.  Causal processing is useful for forming local
tracklets, but ambiguous choices need not be finalized before later evidence arrives.
This motivates a two-stage design: conservative local association followed by global
offline linking.

Large camera motion makes raw image displacement a poor association cue.  The common
background motion must first be estimated and applied to the state prediction.  Long
gaps remain intrinsically ambiguous: a later fragment may be a reappearance or a
different surfer entering after the earlier track ended.  Appearance is therefore
required alongside motion, but the representation should describe the visible sail
rather than a generic pedestrian.

\section{Related Work}

\paragraph{Online tracking by detection.}
SORT combines a Kalman filter with frame-wise assignment and established a strong,
minimal online baseline~\cite{sort}.  Deep SORT adds a learned pedestrian appearance
metric to survive longer occlusions~\cite{deepsort}, while ByteTrack improves
continuity by associating low-confidence detections rather than discarding them
immediately~\cite{bytetrack}.  OC-SORT revisits motion estimation after occlusion and
uses observations to correct accumulated prediction error~\cite{ocsort}.  BoT-SORT
combines motion, camera compensation, and optional appearance features in a stronger
pedestrian tracker~\cite{botsort}; Deep OC-SORT further adapts the contribution of
appearance under feature degradation~\cite{deepocsort}.  These systems motivate our
motion and appearance cues, but remain fundamentally causal at association time.

\paragraph{Offline and global association.}
Batch tracking has long been formulated as network flow or related graph
optimization~\cite{networkflow}.  More recent work learns objectives for multi-frame
flow association rather than training only local pair costs~\cite{globalobjective}.
Our integer program belongs to this family but targets a small, bounded graph of
already conservative tracklets.  It is intentionally simple: start, end, discard,
and link choices are optimized jointly, while the domain-specific work lies in
constructing reliable fragments and costs.

\paragraph{Motion, appearance, and evaluation.}
Kalman filtering~\cite{kalman} and Rauch--Tung--Striebel smoothing~\cite{rts} provide
the temporal state machinery used before and after association.  Learned ReID is
effective when its training semantics match the target class, but pedestrian
embeddings allocate capacity to body and clothing cues that occupy very few pixels in
our footage.  We instead exploit the sail: a large, saturated, deliberately
distinctive surface.  Standard MOT metrics such as HOTA explicitly balance detection,
association, and localization~\cite{hotametric}.  Because our saved observations are
fixed and the product loss is dominated by identity contamination, we report
pair-clustering precision and recall together with absolute contaminated-track and
fragmentation counts rather than claiming an end-to-end MOT score.

\paragraph{Pose-guided framing.}
The stabilization stage is related to pose-guided cropping, but its goal is not human
pose recovery or action recognition.  A windsurf rig is a large articulated object
whose axis-aligned detector box changes strongly with rotation.  Two semantic rig
points are sufficient to construct a rider-relative anchor and scale, making the pose
representation deliberately task-specific.

\section{System Overview}
\label{sec:system}

The pipeline separates perception, identity, and presentation geometry:
\begin{equation}
\begin{aligned}
\text{video} &\rightarrow \text{boxes + pose keypoints}\\
&\rightarrow \text{local tracklets}\\
&\rightarrow \text{offline identity paths}\\
&\rightarrow \text{stable rider views}.
\end{aligned}
\label{eq:system-pipeline}
\end{equation}

The detector is a custom-trained one-class YOLO11m-pose model.  It was initialized
from public pretrained weights and trained for 700 epochs at 640-pixel input
resolution on manually reviewed boxes and two windsurf-specific pose keypoints: the
boom--mast junction and mast tip.  Each box is also converted into a compact sail-color
descriptor after border-connected background suppression.  Independently, background features
are estimated between adjacent frames while masking detected surfers; the resulting
camera transforms are passed to both tracking stages.

The first association stage is causal but deliberately conservative.  It accepts
only locally unambiguous matches and exposes uncertainty as separate tracklets.  The
second stage sees the complete recording, creates a bounded graph of feasible
tracklet continuations, and solves for disjoint paths.  Post-processing smooths linked
boxes and interpolates internal gaps, but retains the distinction between observed
and synthesized detections.  The pose stage then converts every dense identity track
into a frame-indexed virtual-camera signal.

The components have explicit contracts.  Detection proposes observations; local
association should prioritize fragment purity; global association should prefer a
split over implausible identity evidence; and stabilization must never change track
identity.  This modularity is important because only the association contract is
evaluated against alternative trackers, while the final view also depends on
detector and pose-keypoint quality.

\section{Offline Identity Association}
\label{sec:tracking}

The tracker converts detections into high-purity local fragments and then into
globally consistent paths.  The same camera-motion estimates support both stages.

\begin{figure}[t]
  \centering
  \includegraphics[width=\linewidth]{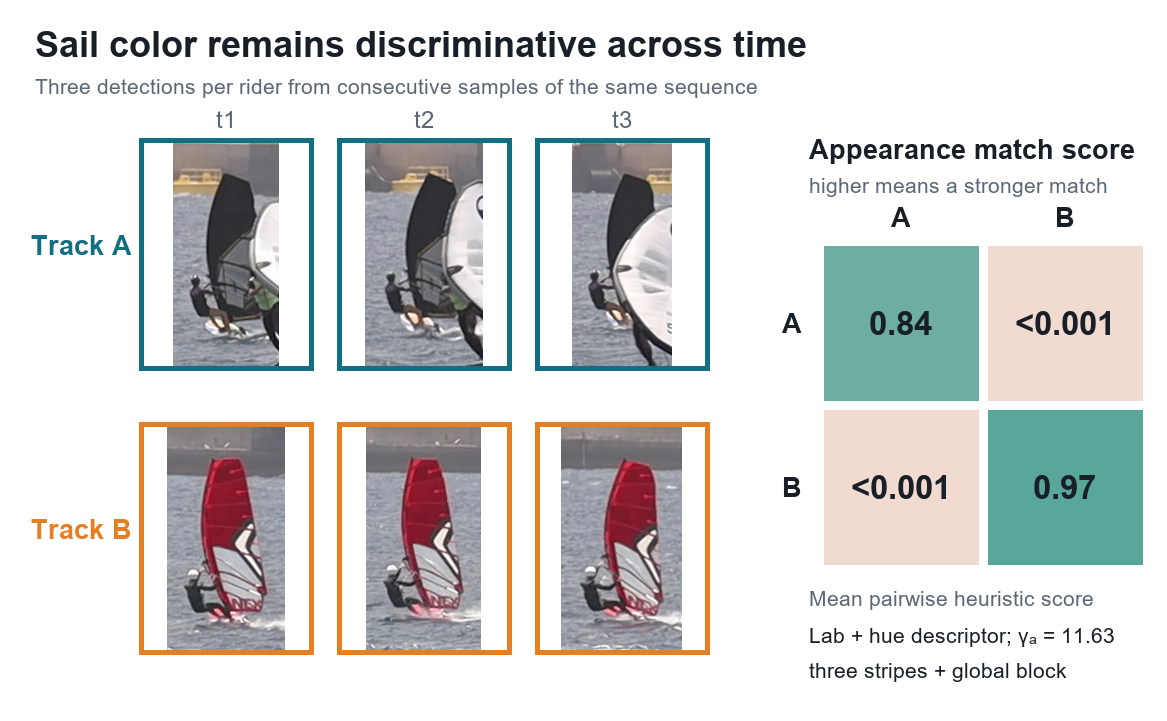}
  \caption{Production color descriptor on consecutive crops of two sails.  Repeated
  observations score highly, while cross-sail matches reach the lower bound.  The
  hand-selected example explains the cue rather than measuring it.}
  \label{fig:tracking-sail-color}
\end{figure}

\subsection{Foreground-masked sail appearance}

Windsurfing sails are large, saturated, and often visually distinctive even when the
rider occupies only a few pixels.  For each detection crop we build a descriptor from
a joint histogram of the chromatic Lab channels $(a,b)$, a circular hue histogram,
and a lower-weight luminance histogram.  Saturation weighting emphasizes colored sail
panels over gray water, spray, and haze.  Three vertical bands preserve coarse color
layout, and a fourth block represents the complete crop.  Each block is L1-normalized
and square-root mapped before the concatenated vector is L2-normalized, yielding a
bounded Hellinger-style distance.

\Cref{fig:tracking-sail-color} illustrates the descriptor response on two visually
distinct sails.

The descriptor first removes likely background using a border-derived Lab model.  A
gradient gate identifies low-texture pixels consistent with that model, and only
candidate background components connected to the crop border are removed.  After
morphological cleanup, only the largest remaining foreground component is retained;
a minimum-mask fallback avoids empty descriptors.  This foreground mask is important: an earlier inactive code
path used a fixed center rectangle, allowing water and overlapping sails to dilute
the intended cue.

\begin{figure*}[t]
  \centering
  \includegraphics[width=0.82\textwidth]{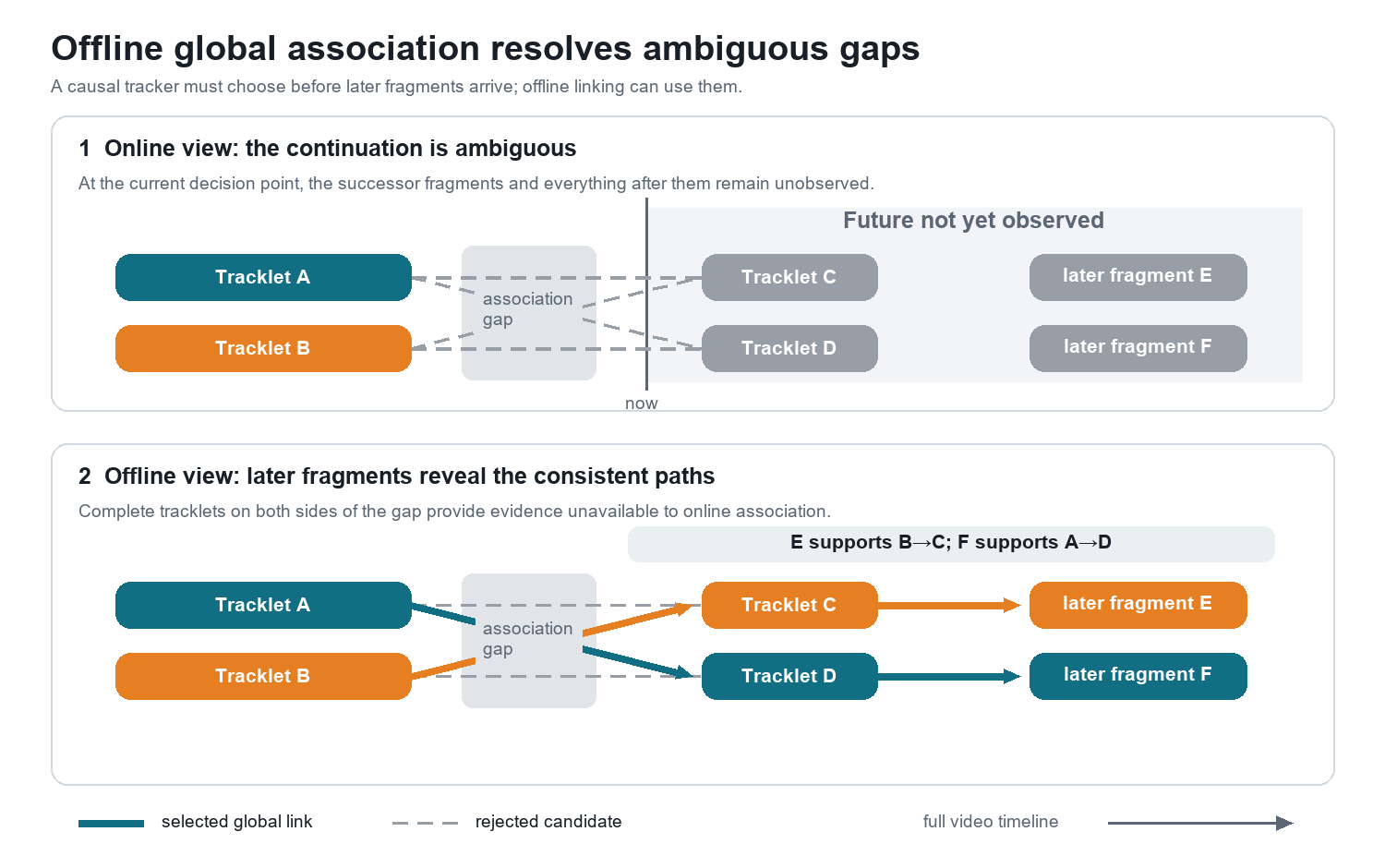}
  \caption{Future fragment evidence supports global tracklet association.  Later
  stable observations distinguish locally ambiguous successors; the optimizer then
  chooses a compatible pairing jointly.  Start, end, discard, and gap costs are
  omitted from the conceptual diagram.}
  \label{fig:tracking-offline-association}
\end{figure*}

\Cref{fig:tracking-offline-association} contrasts the locally ambiguous online view
with the later fragment evidence available to the offline optimizer.

For a tracklet, all observation embeddings are averaged into a whole-fragment
prototype.  This replaced endpoint-biased exponential averages, which emphasized the
last crop of the earlier fragment and first crop of the later fragment---exactly the
frames most likely to contain both sails during a crossing.  Given descriptor distance $d$, the
current heuristic match score is
\begin{equation}
 \begin{aligned}
 \epsilon&=10^{-6},\\
 S_a&=\operatorname{clip}_{[\epsilon,\,1-\epsilon]}(1-\gamma_a d),
 &\gamma_a&=11.63.
 \end{aligned}
 \label{eq:appearance}
\end{equation}
It is a tuned similarity score, not a calibrated posterior probability; the graph
uses $C^{\mathrm{appearance}}=-\log S_a$.  The coefficient $\gamma_a$ was tuned on the
development set.  The numerical margin $\epsilon$ keeps the logarithm well defined.

\subsection{Conservative local tracklets}

Each detection initially forms a singleton.  A greedy preprocessor advances a
constant-velocity Kalman state~\cite{kalman} through the estimated frame-to-frame
camera transforms and compares the prediction with current detections.  Strict links
require both motion and appearance evidence; looser links are accepted only when they
are unique.  Duplicate proposals, competing successors, or multiple plausible
candidates remain separate.  The local stage therefore sacrifices completeness to
preserve fragment purity.

This behavior is deliberate.  In the evaluation reconstruction, 598 of 601
preprocessor fragments are pure with respect to retained gold identities.  The
dominant catastrophic error is introduced later when two pure fragments are linked
incorrectly, which makes the offline edge decision the appropriate place for
additional safeguards.

\subsection{Candidate graph}

For non-overlapping fragments $T_i$ and $T_j$, a directed candidate $i\rightarrow j$
is created only when the gap is at most three seconds.  A Kalman state fitted to all
detections of $T_i$ is propagated to the first observations of $T_j$, applying global
camera transforms at every intervening frame.  Unlike the earlier production call,
the gate uses the complete measurement $(c_x,c_y,w,h)$ rather than discarding width
and height.

Every candidate receives
\begin{equation}
 \begin{aligned}
 C_{ij}={}&w_m C^{\mathrm{motion}}_{ij}
       +w_a C^{\mathrm{appearance}}_{ij}\\
       &+w_g C^{\mathrm{gap}}_{ij},\\
 C^{\mathrm{gap}}_{ij}={}&-\Delta f_{ij}\log p_{\mathrm{miss}}.
 \end{aligned}
 \label{eq:edge-cost}
\end{equation}
For each of the first two observations of $T_j$, the predicted Kalman state gives
squared Mahalanobis distance $d^2$ over $(c_x,c_y,w,h)$.  Motion cost is the mean
negative log chi-square survival score, $-\log\Pr(\chi^2_4\geq d^2)$; appearance uses
the whole-fragment prototypes; and the gap term increases with the number of missing
frames.

The graph now applies absolute plausibility vetoes before optimization.  Edges are
removed when motion cost exceeds 8, or when endpoint box area or aspect ratio changes
by more than a factor of three.  The Kalman filter already models size; the explicit
ratios supplement it with an interpretable discontinuity bound.  These rules implement
abstention at the evidence boundary.  They are necessary because a central fragment's
start and end penalties can otherwise make even poor finite evidence cheaper than
leaving two trajectories separate.

\subsection{Global path optimization}

Binary variables represent starting or ending a path at each fragment, selecting a
candidate edge, or discarding an eligible short fragment.  Each retained fragment has
exactly one incoming choice and one outgoing choice, so selected links form disjoint,
time-ordered paths.  Start and end costs are reduced near image borders, where a
surfer entering or leaving the field of view is more plausible.  An edge that costs
more than ending $T_i$ and starting $T_j$ independently is dominated and removed; the
remaining outgoing degree is bounded before solving the integer program.

The method is offline in a precise sense: it compares stable fragments on both sides
of a gap and chooses all compatible continuations jointly.  It does not infer through
arbitrarily long absence.  Gaps beyond the horizon remain fragmented, and ambiguous
evidence should do the same.

\subsection{Dense trajectories}

Linked detections remain sparse.  The post-processor removes tracks that are too short
or sparsely observed, performs a forward camera-compensated Kalman pass and backward
Rauch--Tung--Striebel smoothing~\cite{rts}, and interpolates internal gaps.  The dense
trajectory records whether each observation was detected or synthesized, then feeds
the rider-relative stabilization stage.

\section{Pose-Guided Stabilization}
\label{sec:pose}

Tracking determines \emph{which} surfer is present; the downstream view must decide
where to place a virtual camera and how tightly to crop.  An axis-aligned detector box
is a poor control signal because a tall, rotating rig changes box center and height
even when the athlete moves little.  Direct box-based cropping therefore produces
lateral drift and visible zoom pumping.

\subsection{Two task-specific pose keypoints}

The pose model predicts the boom--mast junction and mast tip.  These keypoints were
chosen for rendering geometry rather than as a reduced human skeleton.  The boom
junction provides a repeatable reference near the rider; the boom-to-tip distance
provides a rig-relative scale cue that is less sensitive to in-plane rotation than box
height.  The checked-in development dataset contains 1,123 manually labeled training
images and 56 validation images.  The split is image-level and can contain frames from
the same source sequence on both sides, so it does not establish cross-sequence
generalization.

\begin{figure}[t]
  \centering
  \includegraphics[width=0.92\linewidth]{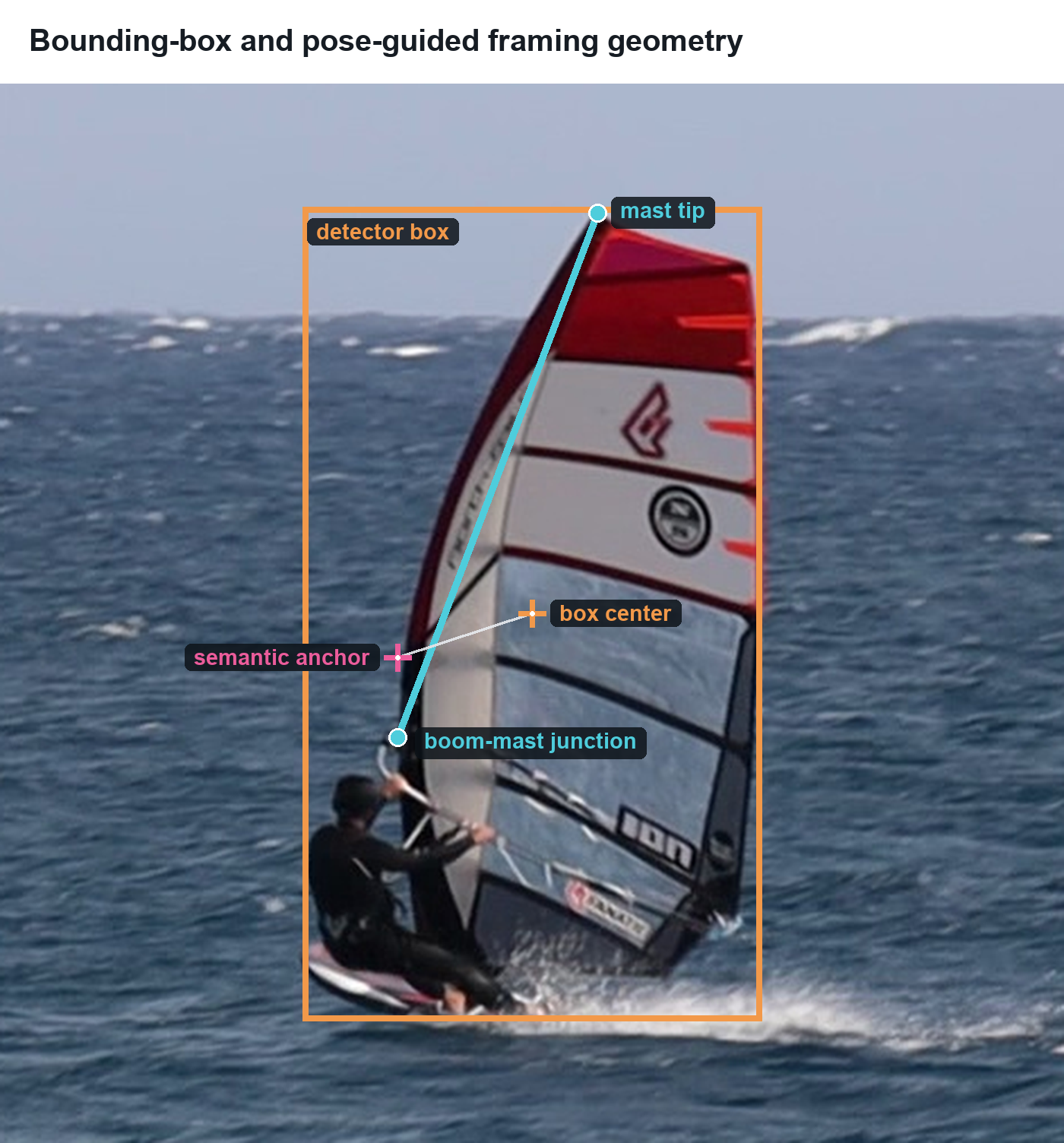}
  \caption{Pose-derived virtual-camera geometry.  The detector box and center
  (orange) describe image extent.  Boom junction and mast tip (cyan) provide a
  rider-relative anchor (magenta) and scale cue.}
  \label{fig:pose-anchor-geometry}
\end{figure}

\Cref{fig:pose-anchor-geometry} summarizes the detector geometry, the two predicted
keypoints, and the derived anchor and scale signals used by the virtual camera.  The
figure makes the distinction visible: the box center follows the axis-aligned extent
of the complete rig, whereas the pose-derived anchor remains tied to the boom--mast
region near the rider.  As the sail rotates, the box center can move even when the
rider moves little; using the pose-derived point therefore reduces this geometric
source of lateral drift before temporal smoothing.

For box top $y_1$, height $h$, and center $(c_x,c_y)$, the fallback anchor is
\begin{equation}
 \mathbf a_{\mathrm{box}}=(c_x,y_1+0.6h).
\end{equation}
When the boom junction $\mathbf b=(b_x,b_y)$ is visible, a low-noise mast proxy
$\tilde{\mathbf t}=(b_x,y_1)$ gives
\begin{equation}
 \mathbf a_{\mathrm{pose}}=0.15\tilde{\mathbf t}+0.85\mathbf b.
 \label{eq:pose-anchor}
\end{equation}
The measured mast tip is intentionally excluded from position: its long lever arm
amplifies localization noise.  It remains useful for scale.

Small crops do not support equally reliable keypoints.  The unsmoothed anchor blends
pose and box geometry according to the shorter box dimension, with pose weight rising
from zero at 45 pixels to one at 55 pixels.  Missing boom observations are interpolated
for short gaps; longer gaps ease toward the per-frame box fallback and back to pose
when reliable evidence returns.  A five-frame robust weighted mean suppresses
isolated jumps while preserving response to actual motion.

\subsection{Rig-relative scale}

When both keypoints are visible, scale begins with
$\ell=\lVert\mathbf b-\mathbf t\rVert_2$.  Short gaps are interpolated or held;
otherwise the anchor-to-box-top distance supplies a fallback.  A centered 31-frame
mean (radius 15) produces $\bar\ell$, and the requested crop height is
\begin{equation}
 C=\max\!\left(\frac{\bar\ell}{0.75},\frac{h}{0.9},1\right),
 \qquad S_m=\operatorname{clip}_{[\epsilon,\,1]}\!\left(\frac{C}{H}\right),
 \label{eq:pose-scale}
\end{equation}
where $H$ is source height and $S_m$ is the mast-relative crop scale.  The mast segment is targeted to 75\% of the output, while
the box term prevents excessively tight crops.  The resulting anchor and scale are
stored per frame with the identity track.  Global background stabilization remains a
separate signal: it improves motion estimation and optional overview rendering, but
cannot replace rider-relative composition when the rig rotates within the scene.

\section{Evaluation}
\label{sec:evaluation}

We evaluate identity association on 21 manually curated development videos.  The
benchmark fixes the universe of 41,079 saved, non-interpolated detections and gives
every method the same bounding boxes.  Of these, 41,004 belong to retained surfer
identities and 75 are annotated for rejection.  This deliberately isolates
association from detector recall: the results answer whether the available
observations were grouped correctly, not whether YOLO found every surfer.

The saved observations were selected during manual trajectory reconstruction and
their boxes were Rauch--Tung--Striebel smoothed within gold identity trajectories.
This is stronger leakage than merely conditioning on useful boxes: the motion cue
receives geometry derived from the target identity partition.  The same videos also
informed system design, and the provenance of the original ILP parameter tuning is
not fully recorded.  The benchmark is therefore an in-sample engineering diagnostic
and failure analysis, not a clean end-to-end or held-out tracker comparison.

\subsection{Metrics}

For retained detections, pairwise precision and recall compare every unordered pair.
A pair is positive when both observations have the same annotated identity:

\begin{align}
  P_{\mathrm{pair}} &=
    \frac{\text{correct same-identity pairs}}
         {\text{all pairs joined by the tracker}}, \\
  R_{\mathrm{pair}} &=
    \frac{\text{correct same-identity pairs}}
         {\text{all annotated same-identity pairs}}.
\end{align}

False merges reduce precision and fragmentation reduces recall.  Missing outputs are
treated as singleton predictions rather than being silently omitted.  Micro scores
pool all pairs and consequently weight long identities quadratically; macro scores
average per-video values.  Coverage is the fraction of retained observations emitted
by a method.  A contaminated predicted track contains more than one gold identity;
``contaminated observations'' counts every observation in such a track and therefore
does not distinguish a brief intrusion from a wholly wrong trajectory.  If $k_g$ is
the number of predicted pieces containing observations of gold identity $g$, then
fragmentation excess is $\sum_g\max(k_g-1,0)$, including pieces shared through a
contaminated prediction.  An exact video has full coverage and a partition identical
to gold up to track-label permutation.

\subsection{Fixed-observation comparison}

\Cref{tab:association-benchmark} compares the production pipeline with OC-SORT and
BoT-SORT.  OC-SORT receives the fixed detections directly.  BoT-SORT uses its ECC
camera compensation but no pedestrian-specific ReID network.  The production method
reruns conservative preprocessing, masked video stabilization, and offline ILP
association.  Its sail descriptor uses border-derived foreground masking and a
whole-fragment appearance prototype; motion gating evaluates center, width, and
height; and implausible motion, area, or aspect-ratio transitions are excluded before
optimization.

\begin{table*}[t]
  \centering
  \footnotesize
  \caption{Fixed-observation identity-association results on 21 development videos.
  ``Contam.'' is contaminated tracks over emitted tracks; ``C. obs.'' counts all
  observations contained in those tracks.}
  \label{tab:association-benchmark}
  \begin{tabular}{lrrrrrrrr}
    \toprule
    Method & Coverage & Pair P & Pair R & Pair F1 & Macro F1 & Contam. & C. obs. & Frag. excess \\
    \midrule
    OC-SORT & 0.981 & \textbf{0.985} & 0.662 & 0.792 & 0.857 & 9/165 (5.5\%) & 6,184 & 845 \\
    BoT-SORT, no ReID & 0.996 & 0.920 & 0.752 & 0.828 & 0.858 & 16/142 (11.3\%) & 14,459 & 244 \\
    Production offline ILP & \textbf{0.999} & 0.957 & \textbf{0.918} & \textbf{0.937} & \textbf{0.970} & 9/95 (9.5\%) & 9,982 & \textbf{42} \\
    \bottomrule
  \end{tabular}
\end{table*}

Production reduces fragmentation from 845 to 42 relative to OC-SORT while retaining
the same absolute number of contaminated tracks.  This continuity gain comes with
lower pair precision, a larger contaminated-output fraction (9/95 versus 9/165), and
more observations contained in contaminated tracks (9,982 versus 6,184).  Eleven of
21 videos are exact, compared with four for BoT-SORT and one for OC-SORT.  The result
is the selected operating point for this application, but not a uniform
improvement and not the nominal zero-false-merge target.  OC-SORT's lower mixed-track
rate comes with 165 outputs for 88 identities, 845 excess fragments, and 1.9\% of
retained observations left unassigned.

A 10,000-sample bootstrap that resamples complete videos gives 95\% intervals for
micro pair F1 of [0.721, 0.864] for OC-SORT, [0.759, 0.909] for BoT-SORT, and
[0.852, 0.995] for production.  The broad, overlapping intervals reinforce the
development-set interpretation rather than a general ranking claim.  The resampling
procedure and seed are recorded with the evaluation artifacts~\cite{projectresources}.

\subsection{Safeguard ablation}

The initial production reconstruction exposed a concentrated false-link problem.
Preprocessor fragments were almost always identity-pure, but the ILP sometimes joined
two pure fragments from different surfers.  Many of the unexpected short-gap errors
occurred at crossings, where the endpoint crops contained both sails.  The previous
appearance model emphasized precisely those boundary frames, the color descriptor's
intended foreground mask was bypassed by an early return, and the ILP requested a
position-only projection from a Kalman model that already represented width and
height.

\Cref{tab:safeguard-ablation} records two staged revisions.  Because several changes
are bundled in the first stage, this is a system ablation rather than a claim that any
single component produced the full difference.  Foreground masking, whole-fragment
appearance aggregation, and four-dimensional motion evaluation raise micro F1 from
0.893 to 0.928.  Adding conservative vetoes for motion cost and threefold area or
aspect-ratio discontinuities raises precision further and halves contaminated tracks,
without increasing fragmentation.

\begin{table*}[t]
  \centering
  \footnotesize
  \caption{Staged production-pipeline revision on the same development set.}
  \label{tab:safeguard-ablation}
  \begin{tabular}{lrrrrr}
    \toprule
    Configuration & Pair P & Pair R & Pair F1 & Contaminated tracks & Fragmentation excess \\
    \midrule
    Original reconstruction & 0.907 & 0.880 & 0.893 & 20 & 48 \\
    + foreground mask, fragment mean, 4D motion & 0.939 & 0.917 & 0.928 & 17 & 44 \\
    + conservative link vetoes & \textbf{0.957} & \textbf{0.918} & \textbf{0.937} & \textbf{9} & \textbf{42} \\
    \bottomrule
  \end{tabular}
\end{table*}

The vetoes are not a second tracker layered on top of the Kalman filter.  The Kalman
state already contains center, width, and height; the earlier ILP call discarded the
size dimensions.  The revised system uses the full four-dimensional Mahalanobis
evidence and additionally refuses links whose absolute evidence is implausible.  This
second step is necessary because the ILP's start and end penalties can otherwise make
even a very poor link cheaper than splitting two central fragments.

A manual audit found nine wrong selected fragment joins across six videos.  This audit
counts only joins between two individually pure preprocessor fragments.  Contaminated
production outputs span seven videos because evaluation sequence 15 instead contains an
already impure local fragment (225 observations from one identity and one from
another), so its adjacent joins are excluded from the pure-to-pure edge count.  Unlike the removed
geometric failures, they concentrate at crossings, visually near-identical sails,
and crops containing both competitors.  Their implications and the unsuccessful
global-margin abstention experiment are discussed in \Cref{sec:discussion}.

\section{Discussion}
\label{sec:discussion}

\subsection{What the comparison establishes}

The benchmark quantifies the intended trade-off.  OC-SORT has the highest pair
precision and nine identity-mixed outputs, but 845 excess fragments make its result
operationally unsuitable.  Production also has nine identity-mixed outputs---a larger
fraction of its 95 outputs---while reducing fragmentation to 42.  Aggregate F1 is
therefore a diagnostic rather than an acceptance criterion; the product still
requires complete-trajectory review.

\subsection{Residual failure mode}

The original failure audit separated two kinds of false link.  Geometrically
implausible joins resulted from bypassed foreground masking, endpoint-biased
appearance aggregation, and position-only use of a size-aware motion model.  The
revised evidence and conservative vetoes remove many of these cases.  The remaining
wrong joins cluster at crossings, visually near-identical sails, and crops containing
both competitors.  In those cases, time, position, scale, and color can all support
the wrong continuation.

This residual mode is not well addressed by another global scalar threshold.  An
abstention experiment removed each chosen edge in turn and compared the alternative
objective.  The resulting margin did not separate wrong from correct links: useful
thresholds rejected substantially more correct associations than errors.  A credible
next step would instead model the interaction explicitly---for example, by detecting
paired crossings and comparing pre- and post-overlap appearance jointly.  We leave
that extension unimplemented rather than tuning a development-set-specific margin.

\subsection{Scope and limitations}

The videos are a development set, and identity-conditioned box smoothing leaks
information into motion; this is neither an end-to-end detector test nor a clean MOTA
or HOTA comparison.  BoT-SORT omits its pedestrian ReID model, while pose-guided
framing has no independent cross-video stability benchmark.  Its image-level split
may also contain nearby frames from the same source video.

The three-second horizon limits false long-range joins and graph size, but longer
genuine disappearances remain split.  The system therefore combines conservative
automation with review rather than guaranteeing perfect recovery.

\section{Conclusion}
\label{sec:conclusion}

We presented an offline system that turns long-shot windsurfing video into stable,
rider-relative trajectories.  Conservative local tracklets, camera-compensated
four-dimensional motion, sail-specific color, global path optimization, and two rig
pose keypoints implement a common design: preserve identity under ambiguity and use
future fragment evidence before committing to longer associations.

On the conditional development-set protocol, production reduces fragmentation from
845 to 42 relative to OC-SORT while retaining nine identity-mixed tracks.  The staged
safeguards reduce identity-mixed tracks from 20 to nine.  This is not a general tracker
ranking; it shows how domain structure, offline inference, and an asymmetric failure
cost can turn custom pose detections into useful rider-relative views.

\clearpage
\appendix
\balance
\section{Reconstruction Details}
\label{sec:appendix}

\subsection{Evaluation reconstruction}

The benchmark uses 21 manually reconstructed development videos.  It contains 41,079 non-interpolated saved
observations: 41,004 assigned to 88 retained identities and 75 assigned to the
discard class.  Saved boxes were RTS-smoothed after manual identity reconstruction.
For the production method, the historical intermediate tracklets were unavailable;
the evaluator therefore starts from singleton saved observations, reruns the current
preprocessor, and then runs the current ILP tracker.  A missing method output is
scored as a unique singleton rather than omitted from the pair metrics.

Predictions were generated with implementation commit
\texttt{a76c2bd774f739e40e1f15a80574ca99bb34d236}.  The final repository snapshot,
aggregate JSON, per-video CSV, protocol notes, and bootstrap script are published as
the project and evaluation artifacts~\cite{projectresources}.
The annotated videos and gold reconstruction files are not distributed, so these
files form an auditable result snapshot rather than an independently rerunnable
artifact.  BoxMOT distribution 13.0.17 was used; its package-level version string
reports 13.0.16.

Runtime is retained only as an operational diagnostic.  On the evaluation machine,
OC-SORT took 9.0 seconds, BoT-SORT 497.7 seconds, and the production association path
402.3 seconds.  These are not comparable speed measurements: OC-SORT needs no video
decode, BoT-SORT includes decode and ECC, and production includes masked stabilization
but excludes initial crop-embedding extraction.

\subsection{Production configuration}

The reported production run uses a maximum candidate gap of three seconds and at
most ten outgoing candidate links per fragment.  The ILP start and end costs are both
96.0993.  Motion, appearance, and gap costs have weights 3.4694, 2.2492, and 9.9926;
the per-frame miss probability is 0.8037 and the appearance coefficient $\gamma_a$ is
11.6301.  The implementation uses the numerical probability margin
$\epsilon=10^{-6}$ for appearance-score clipping.  Motion compares the first two
detections of a successor using the complete $(c_x,c_y,w,h)$ measurement.  Candidate
links are rejected when the motion cost exceeds 8 or when the area or aspect-ratio
change factor exceeds 3.

The appearance prototype retains the full fragment.  Tracklets with at most nine
detections may be discarded, with first-detection cost 36.4497 and growth factor
1.6423.  These values are reported for reconstruction, not as independently
validated universal hyperparameters.

For normalized fragment-end center $(x,y)$ and border margin $m=0.1$, define
$f=\min(1,\max(f_x,f_y))$, where
$f_x=\max(0,m-x,x-(1-m))/m$ and likewise for $f_y$.  Start and end costs are
$w(1-f/2)$: full cost in the central 80\% of the image and half cost at the border.
Sensitivity to the three-second horizon and the motion, area, and aspect-ratio veto
thresholds was not measured.

\subsection{Baseline configuration}

OC-SORT uses the frozen BoxMOT defaults recorded by the evaluator: confidence
thresholds 0.1 and 0.2, maximum age 30, minimum hits 3, IoU threshold 0.3,
$\Delta t=3$, inertia 0.2, BYTE association disabled, $Q_{xy}=0.01$, and
$Q_s=0.0001$.  BoT-SORT uses ECC global-motion compensation with ReID disabled;
its high, low, and new-track thresholds are 0.5, 0.1, and 0.6, its buffer is 30,
its match threshold is 0.8, and score fusion is disabled.  All methods receive the
same fixed observations.

\subsection{Detector checkpoint}

The supporting detector is YOLO11m-pose initialized from
\texttt{yolo11m-pose.pt} and trained for 700 epochs at $640\times640$ input
resolution with deterministic seed 0.  The released checkpoint is
\path{video_processing/inference/weights/yolo_models/windsurfing_pose/best.pt},
with SHA-256
\nolinkurl{4c475caeba9ede3cfc56277161284a4a703be9ff46ae8735a788aac093ca9369}.
The image-level development split
contains 1,123 training and 56 validation images.

\subsection{Audited wrong links}

The final link audit found nine wrong selected links in six videos: one each in
evaluation sequences 12, 16, and 18, and two each in sequences 19--21.  These neutral
indices follow the sequence order in the checked-in per-video result file.  This count concerns selected fragment edges and
is distinct from the number of contaminated output tracks or observations.  In
particular, the nine contaminated outputs span seven videos: the edge audit counts
only wrong links between pure fragments.  The seventh sequence, evaluation sequence 15,
instead contains one impure local fragment with a 225-to-1 identity mixture and is
therefore excluded from that classification.

\end{document}